\documentclass[conference]{IEEEtran}
\IEEEoverridecommandlockouts
\usepackage{cite}
\usepackage{amsmath,amssymb,amsfonts}
\usepackage{algorithmic}
\usepackage{graphicx}
\usepackage{textcomp}
\usepackage{xcolor}
\def\BibTeX{{\rm B\kern-.05em{\sc i\kern-.025em b}\kern-.08em
    T\kern-.1667em\lower.7ex\hbox{E}\kern-.125emX}}

\usepackage{hyperref}
    
\begin{document}

\title{UIESNN: A Scale-Aware Spiking Network for Underwater Image Enhancement\\
}


\author{
\IEEEauthorblockN{
Shuang Chen\IEEEauthorrefmark{1},
Ruochen Li\IEEEauthorrefmark{1},
Zihan Zhu\IEEEauthorrefmark{3},
Ronald Thenius\IEEEauthorrefmark{2},
Farshad Arvin\IEEEauthorrefmark{1},
Amir Atapour-Abarghouei\IEEEauthorrefmark{1}
}
\IEEEauthorblockA{
\IEEEauthorrefmark{1}Computer Science Department, Durham University, UK\\
\{shuang.chen, ruochen.li, farshad.arvin, amir.atapour-abarghouei\}@durham.ac.uk
}
\IEEEauthorblockA{
\IEEEauthorrefmark{2}Institute of Biology, University of Graz,
Graz, Austria.\hspace{5mm}
ronald.thenius@gmail.com
}
\IEEEauthorblockA{
\IEEEauthorrefmark{3}University of Cambridge\hspace{5mm}
zz566@cam.ac.uk
}
}


\maketitle

\begin{abstract}
Underwater image enhancement (UIE) is a practically important yet underexplored application of spiking neural networks (SNNs), where the dominant degradations are large-scale and low-frequency, such as wavelength-dependent colour casts and scattering-induced veiling. Existing SNN restoration designs rely on locally bounded spiking perception, which can limit global correction and lead to saturated or inconsistent representations. To address these challenges, we propose a scale-aware SNN framework for UIE named \textbf{UIESNN}. At its core is a \textbf{Multi-scale Pooling LIF Block} (MPLB) that injects hierarchical multi-scale pooling responses into membrane dynamics, thereby enlarging the effective receptive field while preserving fine-grained details and inducing heterogeneous scale-dependent activations. Building on MPLB, we design a spiking residual architecture that integrates frequency decomposition and attention-based refinement in a fully spike-driven pipeline. Extensive experiments on the EUVP and LSUI benchmarks demonstrate that UIESNN achieves state-of-the-art performance among SNN-based methods, delivering improved colour fidelity and spatial coherence with competitive energy cost.
\end{abstract}


\section{Introduction}

Spiking Neural Networks (SNNs) are widely regarded as energy-efficient and biologically plausible alternatives to conventional deep neural networks. In particular, the Leaky Integrate-and-Fire (LIF) spiking neuron~\cite{maass1997networks,WANG2026108253} integrates input evidence over time and emits discrete spikes, which leads to distinctive spatio-temporal processing behaviours that differ fundamentally from convolutional layers that aggregate spatial context via learned filters. Although a performance gap between SNNs and Artificial Neural Networks (ANNs) often still exists, SNNs have achieved impressive results in high-level vision tasks~\cite{patel2021spiking,luo2024integer}. More recently, they have also been explored for low-level image restoration, such as static image deraining, which is attractive for extreme scenarios such as autonomous driving where cameras must operate under rain~\cite{song2024learning}. However, in another extreme setting, deep-sea exploration, underwater image enhancement (UIE) remains largely underexplored in the SNN literature.

\begin{figure}[t]
    \centering
    \includegraphics[width=0.99\linewidth]{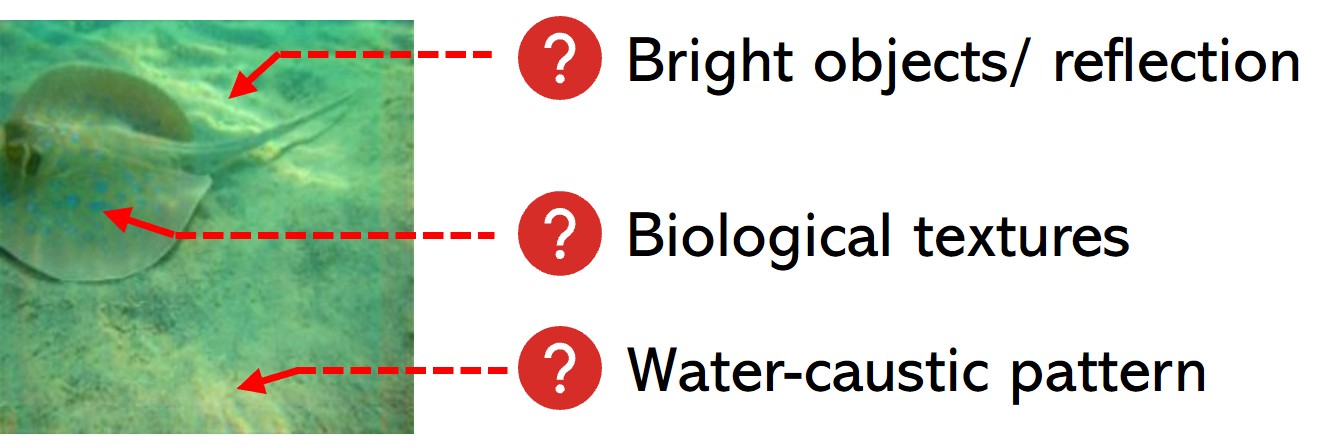}
    \vspace{-0.6cm}
    \caption{Visualisation of complex degradation in the underwater scenario.}
    \vspace{-0.75cm}
    \label{fig:teaser}
\end{figure}

In single-image deraining, the primary challenge is to remove thin, high-frequency rain streaks, whereas UIE focuses on correcting large-scale, low-frequency degradations such as colour casts and haze-like blur caused by light scattering (shown in Fig.~\ref{fig:teaser}). This spectral discrepancy implies that neuron and architecture designs effective for deraining may not generalise to UIE. As shown in~\cite{chen2025exploring}, LIF neurons exhibit task-driven frequency selectivity and behave as high-frequency indicators in deraining. In contrast, UIE requires modelling spatially extensive, slowly varying distortions whose cues are distributed over broad regions. Consequently, a limited, locally bounded receptive field can degrade the ability of spiking neurons to capture the global low-frequency structure underlying underwater degradations. This motivates us to explicitly expand the receptive field in a scale-aware manner.

\begin{figure*}
    \centering
    \includegraphics[width=0.99\linewidth]{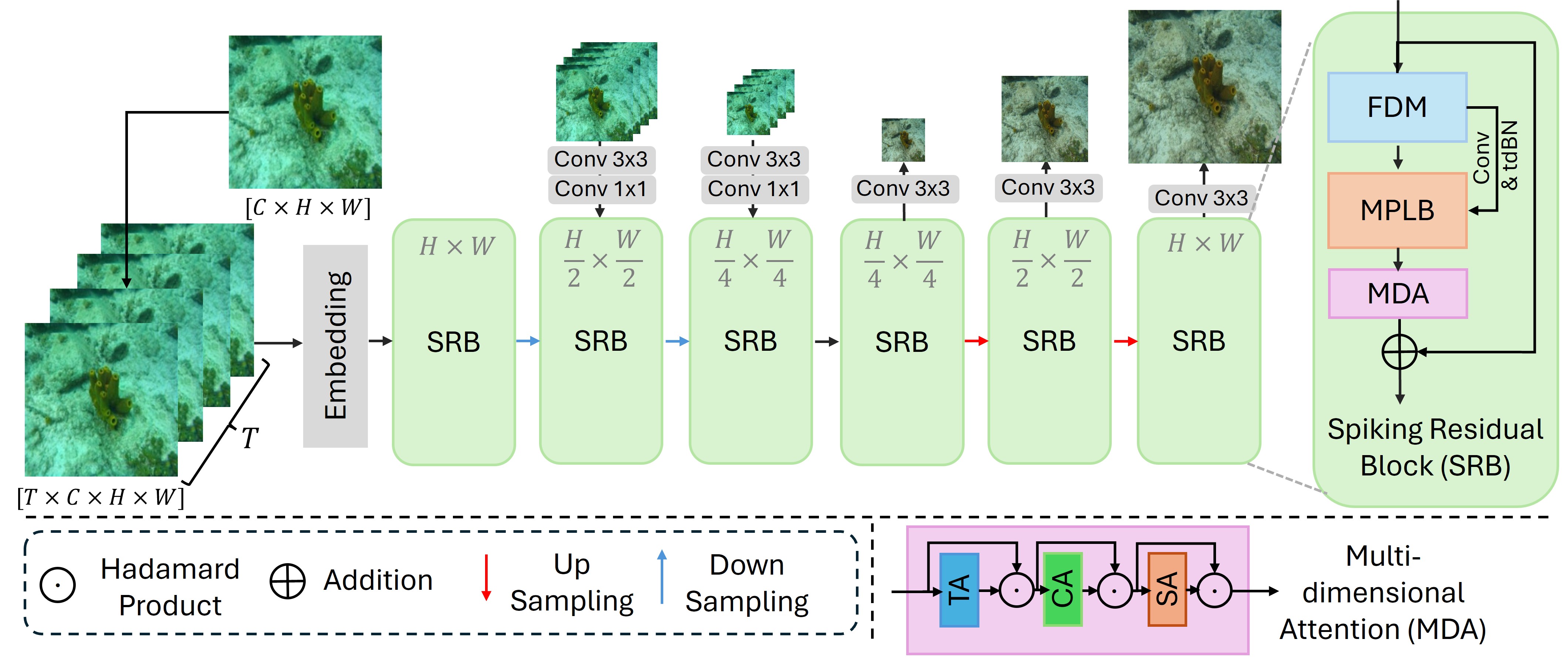}
    \vspace{-0.3cm}
    \caption{Overview of the proposed framework. The top panel presents the overall pipeline. The right panel illustrates the proposed Spiking Residual Network (SRB) for feature refinement with spiking dynamics.}
    \vspace{-0.65cm}
    \label{fig:overview}
\end{figure*}

Another major obstacle lies in the limited spatial perception of conventional spiking neurons. Standard LIF units make firing decisions in a largely point-wise manner, integrating inputs over time but with no explicit mechanism to aggregate wider spatial context. Such spatial insensitivity is particularly restrictive for dense enhancement tasks like UIE, where the degradation cues (e.g., veiling light and global color bias) are distributed over neighborhoods and even image-wide regions rather than isolated pixels. Moreover, as observed in~\cite{chen2025exploring}, na\"{i}vely stacking LIF neurons may lead to frequency-domain saturation, where repeated applications of the same spiking mechanism fail to further enrich the underlying representation. For underwater enhancement, this poses an additional challenge: correcting spatially extensive, low-frequency distortions requires both long-range contextual awareness and the ability to refine representations progressively across scales. These considerations raise a key question: how can we endow spiking neurons with a larger receptive field while preserving fine-grained details, and simultaneously encourage heterogeneous representations that can capture underwater degradations across different spatial extents and frequency bands?

To address this challenge, we propose \textbf{Multi-scale Pooling LIF Block (MPLB)} to expand the receptive field through \emph{hierarchical multi-scale pooling}. Instead of relying on a single fixed neighbourhood, MPLB integrates pooled responses from multiple spatial scales into the membrane potential dynamics, allowing each neuron to jointly encode local details and region-level context. Leveraging the thresholded firing mechanism and task-driven frequency selectivity of spiking neurons, MPLB produces scale-dependent activations: coarse-scale inputs emphasise spatially extensive low-frequency degradations (e.g., veiling light and colour bias), while fine-scale inputs preserve texture-sensitive cues. Therefore, MPLB enlarges spatial awareness without indiscriminately smoothing details, and encourages heterogeneous representations across scales.

Built upon MPLB, we develop \textbf{UIESNN}, an end-to-end SNN framework tailored for underwater image enhancement. UIESNN captures fine-grained structures and global colour/contrast corrections in a spike-driven manner, making it suitable for power-constrained underwater vision systems. Experiments on \textbf{EUVP} and \textbf{LSUI} show UIESNN consistently outperforms prior SNN-based restoration baselines~\cite{song2024learning,chen2025exploring}, which shows importance of scale-aware spiking representations for low-frequency degradations. Code is available: \url{https://github.com/ChrisChen1023/UIESNN}.
Our contributions are:
\begin{itemize}
    \item We identify a core challenge in applying SNNs to UIE: dominant multi-scale low-frequency degradations are insufficiently captured by conventional LIF neurons with limited receptive fields and saturated representations.
    \item We propose \textbf{MPLB}, a multi-scale spike-driven pooling block to enlarge receptive fields while preserving details and inducing heterogeneous feature representations.
    \item We develop \textbf{UIESNN}. Extensive experiments on EUVP and LSUI demonstrate that UIESNN achieves state-of-the-art performance among SNN-based methods.
\end{itemize}


\begin{figure*}
    \centering
    \includegraphics[width=0.99\linewidth]{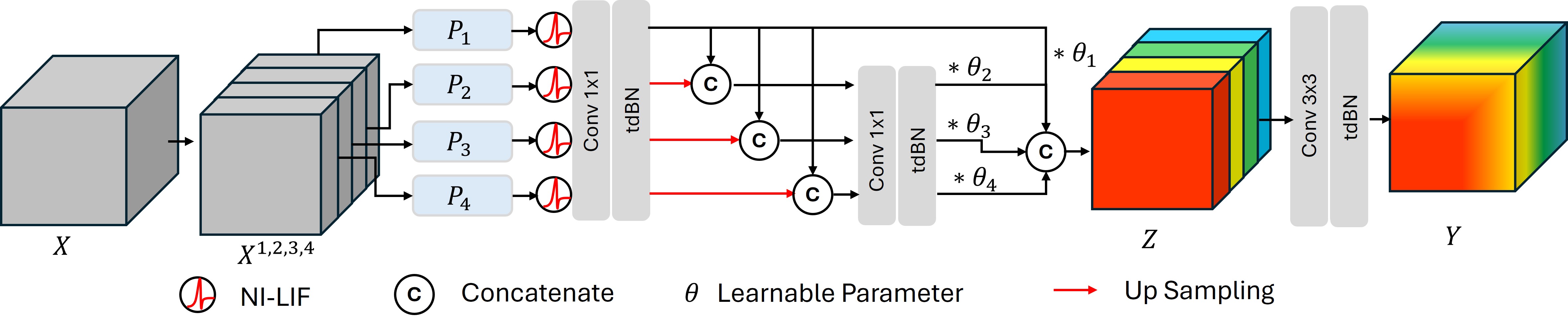}
    \vspace{-0.3cm}
    \caption{The illustration of the proposed Multi-scale Pooling LIF Block. NI-LIF is illustrated in Fig.~\ref{fig:nilif}.}
    \vspace{-0.5cm}
    \label{fig:mplb}
\end{figure*}

\begin{figure}
    \centering
    \includegraphics[width=0.99\linewidth]{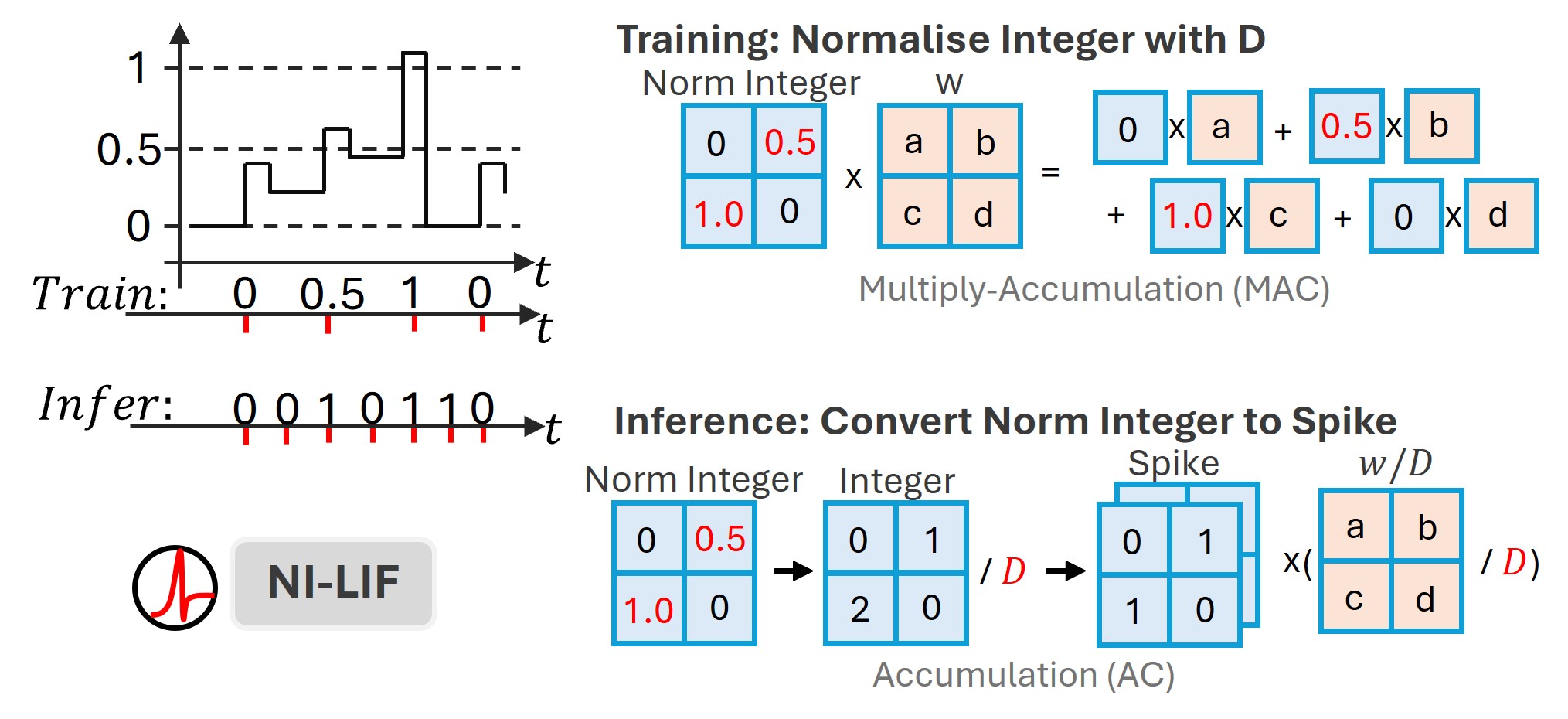}
    \vspace{-0.6cm}
    \caption{The illustration of the NI-LIF~\cite{lei2025spike2former}.}
    \vspace{-0.3cm}
    \label{fig:nilif}
\end{figure}

\section{Related Work}
\subsection{Underwater Image Enhancement}
Underwater image enhancement has progressed from prior-based, handcrafted pipelines to learning-based methods. Early methods relied on imaging priors such as attenuation and scattering, which often generalise poorly across diverse underwater conditions~\cite{chiang2011underwater}. Deep learning has improved robustness by combining data-driven representations with physical insights, including model-based designs~\cite{wang2019underwater}, prior-guided CNNs~\cite{li2020underwater}, and GAN-based restoration~\cite{cong2023pugan}, though GAN training can be unstable and CNN receptive fields remain limited.

To better handle large-scale degradations, recent UIE models adopt Transformers for long-range dependency modeling~\cite{peng2023u} and phase-aware attention for improved structural fidelity~\cite{khan2024phaseformer}. Meanwhile, frequency-domain modelling is increasingly popular, leveraging wavelet or Fourier priors~\cite{zhao2024wavelet}, shallow-layer frequency features~\cite{guo2024underwater}, and spatial–frequency interaction with FFT-based refinement~\cite{chen2025deep} to better capture the low-frequency nature of underwater degradations. 

More recently, energy-efficient SNNs have been explored for UIE. \cite{sudevan2025underwater} proposed a convolutional spiking encoder–decoder, which shows that competitive restoration can be achieved with $T=5$ timesteps while reducing computation and energy consumption. \cite{shao2025lamsnn} further explored SNNs for UIE by directly adopting two existing SNN architectures. However, existing SNN-based UIE efforts have not explicitly addressed the key challenge that arises when spiking neurons are applied to underwater imagery: standard LIF-style dynamics are inherently local and receptive-field limited, and thus struggle to capture multi-level degradations (e.g., region-wide colour casts and haze-like scattering) that dominate underwater scenes. 

\subsection{Spiking Neural Network}
Spiking Neural Networks are increasingly studied as biologically plausible and energy-efficient alternatives to ANNs~\cite{wang2025smformer}, leveraging sparse event-driven spikes for low-power computation~\cite{lei2025spike2former,wang2025spikcommander, wang2025spikcommander}. Most modern SNNs are built on Leaky Integrate-and-Fire neurons and are trained either by converting pre-trained ANNs into spiking counterparts~\cite{cao2015spiking_ann} or by direct training with surrogate gradients~\cite{neftci2019surrogate}. With these advances, SNNs have achieved strong performance on high-level vision tasks such as image classification.

For low-level vision, only a limited number of works have explored SNNs for image restoration~\cite{xu2025snnsir}, and most focus on high-frequency degradations such as single-image deraining~\cite{song2024learning,chen2025exploring}. In contrast, our work targets underwater image enhancement, where degradations are predominantly low-frequency and region-level, and proposes a spiking solution tailored to this distinct challenge.

\begin{figure}
    \centering
    \includegraphics[width=0.99\linewidth]{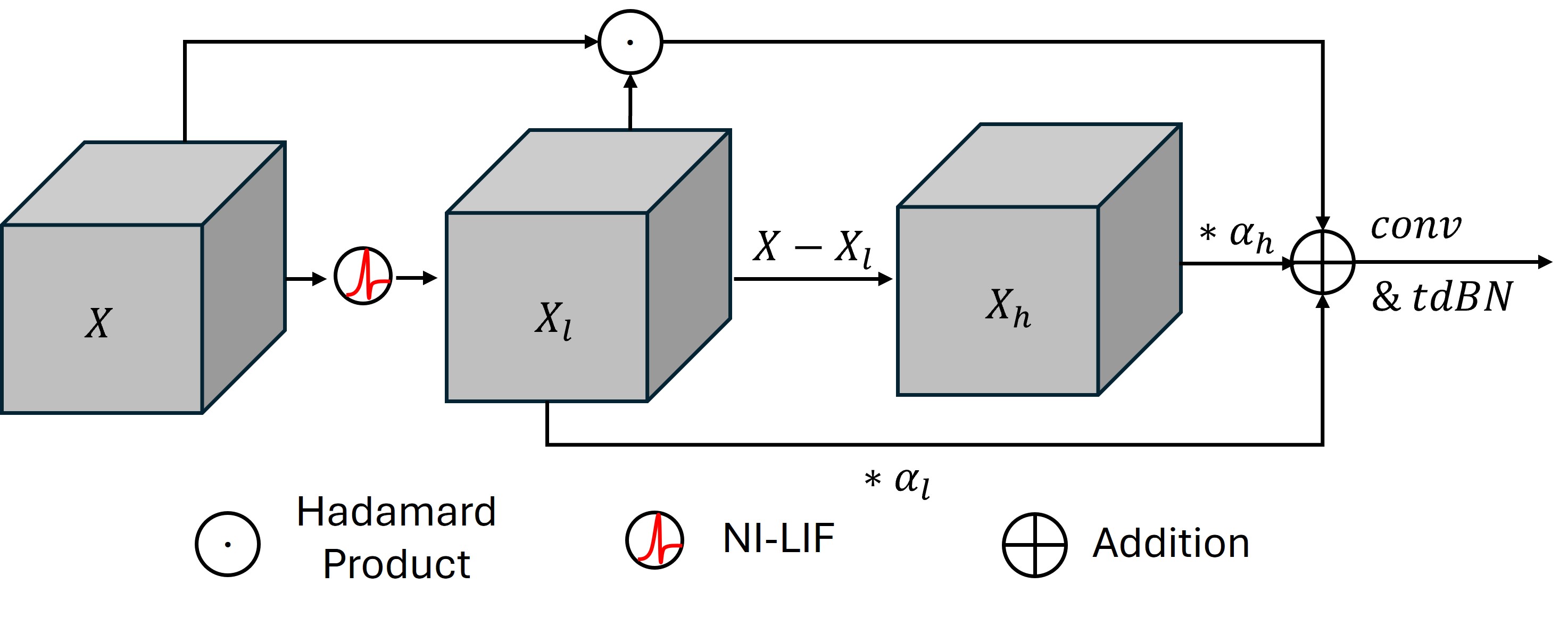}
    \vspace{-0.5cm}
    \caption{The illustration of the frequency decomposition module.}
    \vspace{-0.5cm}
    \label{fig:fdb}
\end{figure}

\section{Methods}
This section presents \textbf{UIESNN} for static underwater image enhancement. We first discuss why spiking neurons require a UIE-specific design, then introduce a \textbf{Multi-Scale Pooling LIF Block} (\ref{sec:method:mplb}) that expands the effective receptive field. Next, we build a \textbf{Spiking Residual Block} (\ref{sec:method:srb}) by integrating frequency decomposition, multi-scale spiking perception, and attention. Finally, we describe the overall encoder--decoder architecture and the multi-scale training objective.

\subsection{Why Spiking Neuron Receptive Fields Matter?}
Underwater image enhancement is dominated by low-frequency degradations such as wavelength-dependent colour casts and scattering-induced veiling. \cite{chen2025exploring} finds that spiking neurons can behave as frequency-domain indicators in a task-driven manner. When the degradation is primarily low-frequency, the spiking responses tend to emphasise low-frequency components. However, a conventional LIF neuron is inherently local because its membrane integrates signals at each spatial location independently, which results in a unit receptive field in the spatial domain. This locality becomes a bottleneck for UIE because underwater degradations are often global yet spatially heterogeneous. As shown in Fig.~\ref{fig:teaser}, a pollution-induced green tint can be global in distribution, but its manifestation becomes non-uniform around bright objects, saturated colors, biological textures, and water-caustic patterns. These heterogeneous low-frequency artifacts frequently appear as irregular regions that are locally continuous, and they are difficult to be reliably captured when the spiking decision is made from a single pixel or a single feature location. This motivates us to design the enlarged receptive fields for spiking neurons to solve this problem.

\begin{figure}
    \centering
    \includegraphics[width=0.99\linewidth]{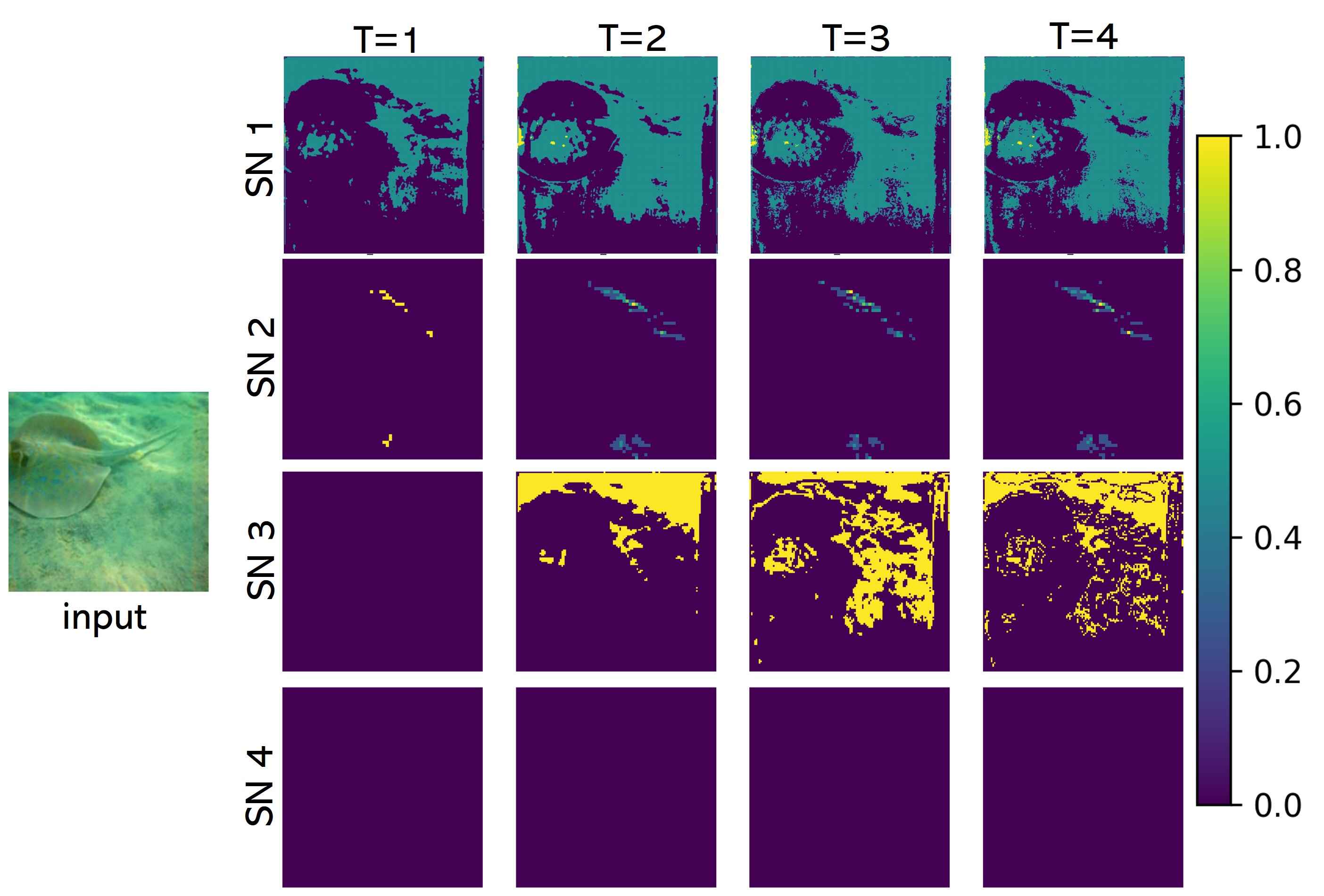}
    \vspace{-0.5cm}
    \caption{Temporal spiking feature visualisation in the Multi-scale Pooling LIF Block. SN1--SN4 denote the four spike neurons in the four scale branches of MPLB after the corresponding $P_{1}$--$P_{4}$ in Sec.~\ref{sec:method:mplb}}
    \vspace{-0.5cm}
    \label{fig:vis-lif}
\end{figure}

\subsection{Multi-scale Pooling LIF Block}
\label{sec:method:mplb}
To address the receptive field limitation without sacrificing efficiency, we propose the
\textbf{Multi-scale Pooling LIF Block} (MPLB). Instead of enlarging receptive fields with
large-kernel convolutions, we use parameter-free average pooling at multiple spatial scales
and let spiking dynamics decide which scale to activate, as illustrated in Fig.~\ref{fig:mplb}. Specifically, given an input tensor
$X \in \mathbb{R}^{T \times B \times C \times H \times W}$, we split it evenly along the channel
dimension into four groups and feed them to four scale branches:
\begin{align}
    \bigl(X^{(1)}, X^{(2)}, &X^{(3)}, X^{(4)}\bigr) = \mathrm{Split}_c(X), \\
X^{(i)} &\in \mathbb{R}^{T \times B \times \frac{C}{4} \times H \times W},
\end{align}
so that $X = [X^{(1)}, X^{(2)}, X^{(3)}, X^{(4)}]_c$ (channel-wise concatenation).
The four branches correspond to: (i) pixel-level perception ($\mathcal{P}_1$), 
(ii) coarse local perception with $4 \times 4$ average pooling, (iii) intermediate local
perception with $2 \times 2$ average pooling, and (iv) global perception with adaptive global
average pooling:
\begin{align}
\mathcal{P}_1&=\mathrm{Identity},\quad \mathcal{P}_2=\mathrm{AvgPool}_{4\times4},\\
\mathcal{P}_3&=\mathrm{AvgPool}_{2\times2}, \quad
\mathcal{P}_4=\mathrm{AdaGAP}.
\end{align}
Each branch applies a spiking neuron, followed by a lightweight $1 \times 1$ convolution and threshold-dependent batch normalisation (tdBN), then upsamples back to the original spatial resolution when pooling is used.
We implement the NI-LIF~\cite{lei2025spike2former} spiking neuron using a normalised multi-spike variant (shown in Fig.~\ref{fig:nilif}). For each branch
$i\in\{1,2,3,4\}$ and time step $t$, we first obtain the input current via the corresponding
pooling operator:
\begin{equation}
U_t^{(i)} = \mathcal{P}_i\!\left(X_t^{(i)}\right),
\end{equation}
then update the membrane potential as:
\begin{equation}
\mu_t^{(i)} = \gamma \bigl(\mu_{t-1}^{(i)} - S_{t-1}^{(i)}\bigr) + U_t^{(i)},
\end{equation}
where $\gamma$ is the decay constant. Inspired by~\cite{lei2025spike2former}, the spike output
is produced by a quantised surrogate activation:
\begin{equation}
S_t^{(i)} = \mathcal{Q}\!\left(M_t^{(i)}\right),
\end{equation}
which clips and rounds membrane values into a small set of discrete spike levels and normalises them to improve stability. After spiking, we apply a projection and normalisation, and upsample if needed:
\begin{equation}
\hat{F}_t^{(i)} = \mathrm{tdBN}\!\left(\mathrm{Conv}_{1\times 1}\!\left(S_t^{(i)}\right)\right), \qquad
F_t^{(i)} = \mathcal{U}_i\!\left(\hat{F}_t^{(i)}\right),
\end{equation}
where $\mathcal{U}_1=\mathrm{Id}$ and $\mathcal{U}_i$ (for $i=2,3,4$) upsamples features back to
$H\times W$. Collecting all time steps yields
\begin{equation}
F_i \in \mathbb{R}^{T \times B \times \frac{C}{4} \times H \times W}, \qquad i=1,2,3,4.
\end{equation}

We then fuse multi-scale features by forming three pairwise mixtures with the pixel-level branch
to enlarge the receptive field while preserving fine-grained details:
\begin{align}
M_1 = \phi\!\bigl([F_1, F_2]_c&\bigr), \quad
M_2 = \phi\!\bigl([F_1, F_3]_c\bigr), \\
M_3 = &\phi\!\bigl([F_1, F_4]_c\bigr),
\end{align}
where $[\,\cdot,\cdot\,]_c$ denotes channel-wise concatenation and $\phi(\cdot)$ denotes a $1 \times 1$
convolution followed by threshold-dependent batch normalisation (optionally projecting back to $\tfrac{C}{4}$ channels for efficiency). In this way, MPLB can respond to low-frequency degradations at multiple spatial supports, from regional bias to global cast, without losing fine-grained detail features.
To enable {input-adaptive} scale selection with negligible overhead, we introduce a set of learnable fusion coefficients ${\theta_i},{i=1,2,3,4}$ to reweight the pixel-level feature and the three mixed features before the final aggregation:
\begin{align}
\tilde{F}_1 &= \theta_1 \odot F_1,\quad \tilde{M}_1 = \theta_2 \odot M_1,\\
\tilde{M}_2 &= \theta_3 \odot M_2,\quad \tilde{M}_3 = \theta_4 \odot M_3,
\end{align}
where $\odot$ denotes element-wise multiplication with $\theta_i$ to introduce an explicit mechanism to resolve cross-scale conflicts: for regions dominated by low-frequency degradations, the model can prioritise the global or coarse pooling. For textures and edges, it can prioritise the pixel-level pathway.
After reweighting, we aggregate all branches via concatenation and a lightweight $3\times 3$ fusion:
\begin{equation}
Z = [\tilde{F}_1, \tilde{M}_1, \tilde{M}_2, \tilde{M}_3],
\end{equation}
\begin{equation}
Y = \mathrm{tdBN}\left(\mathrm{Conv}{3\times 3}(Z)\right),
\end{equation}
where $Y$ is the MPLB output. Notably, the fusion remains efficient: multi-scale context is from parameter-free pooling and small-kernel convolutions, avoiding the expensive cost in large kernel size associated with receptive field expansion. 

\noindent\textbf{Analysis of Multi-scale Pooling LIF Block}
Fig.~\ref{fig:vis-lif} visualises the temporal responses of the four NI-LIF branches in MPLB. The multi-scale pooling operators provide different spatial ranges, therefore each branch produces a distinct spiking pattern. The finest branch (SN 1) shows more spatially selective activations that are helpful for preserving local structures and texture cues, while the pooled branches produce more region-wise responses that better capture the degradations from specific regions, such as global colour cast and veiling light (SN 2). Across timesteps, the post-spike features generally become stronger and more complete because membrane integration accumulates evidence before firing, which helps aggregate weak but globally consistent cues. Moreover, different branches respond with different strengths over time, suggesting that MPLB can emphasise the most informative scales for a given input, and this scale-aware fusion improves colour fidelity and spatial coherence in the restored results.



\subsection{Spiking Residual Block}
\label{sec:method:srb}
Based on MPLB, we design a \textbf{Spiking Residual Block} (SRB).
The block contains two consecutive groups. Inspired by the task-driven frequency indication behaviour~\cite{chen2025exploring}, as shown in Fig.~\ref{fig:fdb}, we build the first group as the Frequency Decomposition Module (FDM), which uses a spiking neuron to perform frequency-like decomposition. Given input $X$, we compute spiking responses:
\begin{equation}
X_l = \mathrm{LIF}(X), \quad X_h = X - X_l ,
\end{equation}
where $X_l$ acts as a low-frequency feature spiked by the indicator LIF~\cite{chen2025exploring} and $X_h$ retains the residual component. We apply learnable scalars $\alpha_l$ and $\alpha_h$ to optimise the two components, and we further introduce an element-wise gated enhancement term $X \odot X_l$. The fused representation becomes
\begin{equation}
\widetilde{X} = \alpha_l X_l + \alpha_h X_h + (X \odot X_l).
\end{equation}
The $\widetilde{X}$ is fed into a combo of $3 \times 3$ convolution and tdBN.
The second group applies MPLB to enlarge receptive fields and capture multi-level low-frequency degradations. Specifically, we compute:
\begin{equation}
Z = \mathrm{MPLB}(\mathrm{tdBN}(\mathrm{Conv}(\widetilde{X}))).
\end{equation}
Then we apply another $3 \times 3$ projection and normalisation. Finally, we combine residual learning and attention. We use a projected shortcut path to match distributions and a Multi-Dimensional Attention~\cite{yao2023attention} (MDA) module to refine the output. The block output is
\begin{equation}
Y = \mathrm{MDA}(Z + \mathrm{Shortcut}(X)) + X.
\end{equation}
By design such a Spiking Residual Block, we preserve stable information flow while allowing the spiking pathway to focus on region-level underwater degradations.

\begin{table*}[t]
\centering
\caption{Quantitative comparison on EUVP and LSUI in terms of PSNR, SSIM, Params (M), and energy (mJ). \textbf{Bold} indicates the best performance in each column, and \underline{underline} indicates the second-best. ESDNet and VLIF are fully trained on EUVP and LSUI following their official settings.}
\resizebox{0.95\textwidth}{!}{
\begin{tabular}{l|ccccc|ccccc}
\hline
\hline
Dataset & \multicolumn{5}{c|}{EUVP} & \multicolumn{5}{c}{LSUI} \\
\cline{2-11}
 & \multicolumn{2}{c|}{ANN}&\multicolumn{3}{c|}{SNN}  & \multicolumn{2}{c|}{ANN}&\multicolumn{3}{c}{SNN}
\\
\cline{2-11}
Metric  & TACL~\cite{liu2022twin} & UIE-WD~\cite{ma2022wavelet}  & ESDNet~\cite{song2024learning} & VLIF~\cite{chen2025exploring} & Ours
        & TACL~\cite{liu2022twin} & UIE-WD~\cite{ma2022wavelet}  & ESDNet~\cite{song2024learning}& VLIF~\cite{chen2025exploring} & Ours \\
\hline
PSNR$\uparrow$      & 20.99 & 17.80   & 25.36 & \underline{25.69}
                  & \textbf{26.97}    & 22.97    &  19.23   & 23.8747  & \underline{24.1731}    & \textbf{24.7346}  \\
SSIM$\uparrow$     & 0.782 & 0.760          & 0.8645
                  & \underline{0.8817}    & \textbf{0.8936}     &  0.8280    &  0.8036      & 0.8725   & \underline{0.8744}  & \textbf{0.8754} \\
Params(M)$\downarrow$   & 28.29 & 14.46          & \textbf{12.81}
                  & \underline{15.72}    & 16.72& 28.29 & 14.46          & \textbf{12.81}
                  & \underline{15.72}    & 16.72  \\
Energy(mJ)$\downarrow$  & 1104.3& 472.7    & \textbf{174.63} 
                  & {697.63}    & \underline{199.33}  & 1104.3& 472.7    & \textbf{174.63} 
                  & {697.63}    & \underline{199.33} \\
\hline
\hline
\end{tabular}
}
\label{tab:comp_euvp_LSUI}
\vspace{-0.6cm}
\end{table*}
\subsection{Overall Architecture and Loss Function}
\textbf{Overall architecture.} UIESNN adopts a three-level encoder-decoder architecture composed of stacked Spiking Residual Blocks. For a static input image $I \in \mathbb{R}^{B \times 3 \times H \times W}$, we replicate it along the temporal axis to obtain $I^{(t)}$ for $t=1,\dots,T$, following the multi-step spiking pipeline. A shallow overlap patch embedding layer maps the input to the feature space using a $3 \times 3$ convolution in multi-step mode. The encoder contains Level 1, Level 2, and Level 3 stages with progressive downsampling. 

To improve robustness to diverse underwater degradations, we inject multi-scale inputs into deeper encoder stages. Specifically, at Level 2 and Level 3, we downsample the original image to the target resolution, embed it using lightweight convolutions, concatenate it with encoder features, and then compress channels with a linear projection. This provides explicit low-level references at multiple scales and helps colour and illumination correction.

The decoder follows a symmetric hierarchy with fewer spiking residual blocks per level. Each level upsamples features per timestep, then fuses the corresponding encoder features via skip connections and refines them with stacked Spiking Residual Blocks. UIESNN outputs predictions at Level 3 and Level 2, and a final full-resolution result at Level 1. For each head, we temporally average features using $\mathrm{mean}$ before a final $3\times3$ reconstruction convolution, and add the input image as a global residual to obtain the restored output.

\textbf{Training objective.} We supervise three outputs using a multi-scale loss. $\hat{I}^{(1)}$ is the final output, $\hat{I}^{(1/2)}$ and $\hat{I}^{(1/4)}$ are the Level 2 and Level 3 outputs after resizing to the full resolution and adding the residual input in the forward pass. During training, we compare each prediction with the ground truth at the corresponding scale by downsampling the target and the prediction with bilinear interpolation.

We define the content loss as an $L_1$ loss summed over scales
\begin{equation}
\mathcal{L}_{\text{pix}} = \sum_{s \in \{1,\,1/2,\,1/4\}} \left\|D_s(\hat{I}^{(s)}) - D_s(I_{\text{gt}}) \right\|_1,
\end{equation}
where $D_s(\cdot)$ denotes bilinear downsampling by scale factor $s$ and $I_{\text{gt}}$ is the ground truth image. We also apply an SSIM loss at each scale
\begin{equation}
\mathcal{L}_{\text{ssim}} = \sum_{s \in \{1,\,1/2,\,1/4\}} \left(1 - \mathrm{SSIM}\left(D_s(\hat{I}^{(s)}), D_s(I_{\text{gt}})\right)\right).
\end{equation}
To explicitly regularise frequency content, we further include a Fourier-domain loss. Let $\mathcal{F}(\cdot)$ denote the 2D FFT. We compute the frequency discrepancy by measuring the $L_1$ distance between the Fourier spectra of the prediction and the ground truth at multiple scales:
\begin{equation}
\mathcal{L}_{\text{fft}} =
\sum_{s \in \{1,\,1/2,\,1/4\}}
\left\|
\mathcal{F}\bigl(D_s(\hat{I}^{(s)})\bigr)
-
\mathcal{F}\bigl(D_s(I_{\text{gt}})\bigr)
\right\|_1.
\end{equation}
The final loss is:
\begin{equation}
\mathcal{L} = \lambda_{\text{pix}} \mathcal{L}_{\text{pix}} + \lambda_{\text{ssim}} \mathcal{L}_{\text{ssim}} + \lambda_{\text{fft}} \mathcal{L}_{\text{fft}},
\end{equation}
where we set $\lambda_{\text{fft}}=0.1$ and $\lambda_{\text{ssim}}=1$, and $\lambda_{\text{pix}}=0.5$. This objective jointly enforces spatial fidelity, structural consistency, and frequency alignment, which is particularly important for underwater restoration where low-frequency color and illumination shifts coexist with fine texture details.

\begin{figure}
    \centering
    \includegraphics[width=0.99\linewidth]{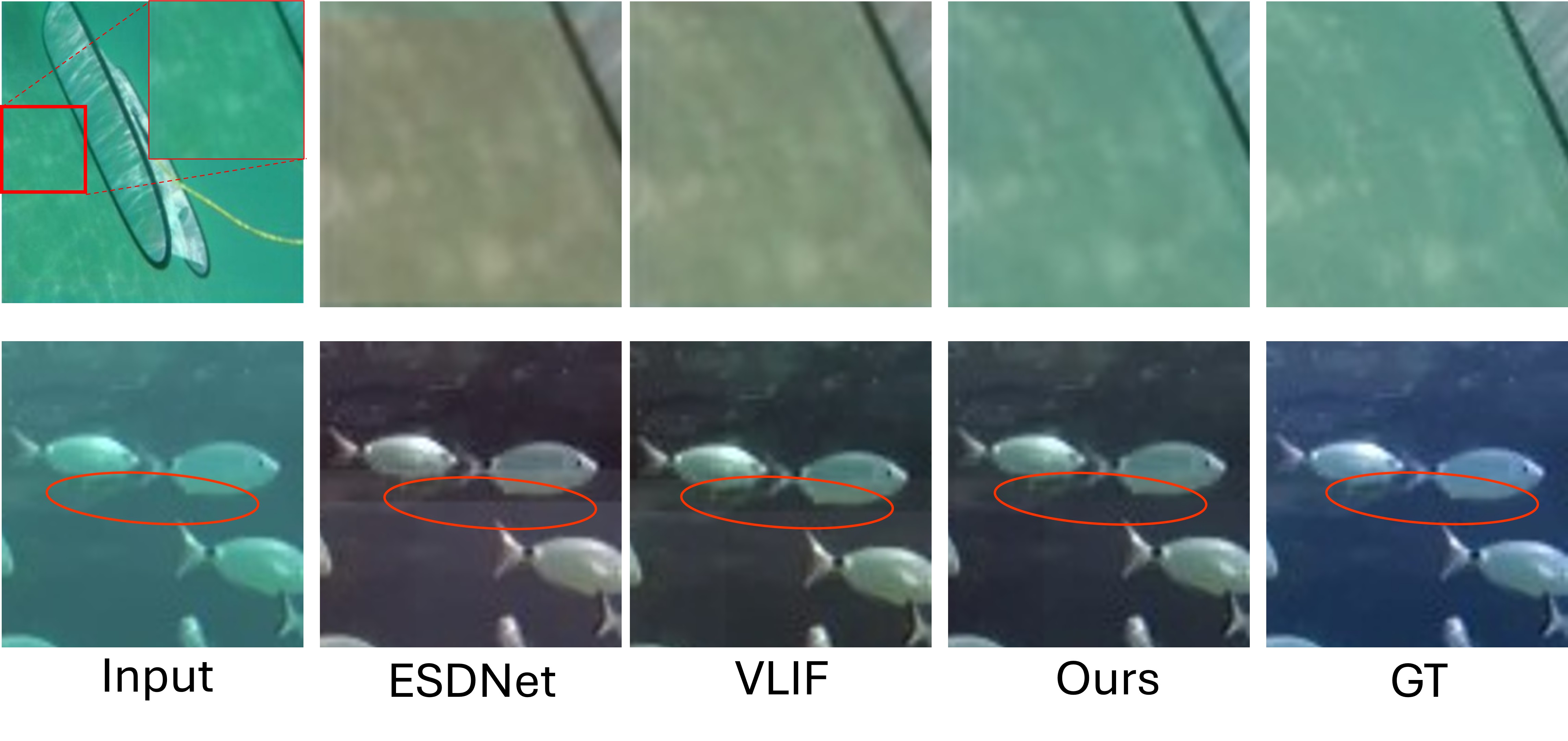}
    \vspace{-0.5cm}
    \caption{Qualitative comparison on underwater image enhancement. The top and bottom rows are results on EUVP and LSUI, respectively. The red boxes (top row) highlight that our method restores more accurate colours and finer texture details, while the red ellipses (lower rows) show that our method produces more spatially coherent structures with fewer discontinuities compared with prior SNN-based methods.}
    \vspace{-0.5cm}
    \label{fig:long}
\end{figure}

\section{Experiments}
\subsection{Setting and Energy Calculation}
\noindent\textbf{Setting}
Unless noted otherwise, we run all comparative experiments under unified training protocols. Our UIESNN is built by stacking Spiking Residual Blocks in a stage-wise layout of $[4,4,8,2,2,2]$. Both downsampling and upsampling are implemented with spike-driven convolutional operators to maintain a fully event-based feature transformation throughout the network. We set the virtual timestep scaling constant to $D{=}4$ and use $T{=}4$ timesteps in all experiments. During training, we crop patches of $64\times 64$ with a batch size of 12. Following~\cite{song2024learning}, we adopt the Sigmoid surrogate function for gradient backpropagation. All models are implemented in PyTorch and trained on a single NVIDIA A6000 GPU. We evaluate UIESNN on the widely used EUVP~\cite{islam2020fast} and LSUI~\cite{peng2023u} benchmarks. Since official checkpoints are not publicly available for the compared SNN baselines, we strictly follow the original experimental configurations of ESDNet and VLIF to reproduce their performance on UIE. Unless specified otherwise, ablation studies are conducted on EUVP.

\noindent\textbf{Energy Calculation}
We estimate inference energy using an operation-count proxy that is widely adopted in the SNN literature~\cite{yao2023attention,luo2024integer,lei2025spike2former}, where the dominant cost is attributed to multiply-accumulate (MAC) operations in ANNs and spike-triggered accumulation (AC) operations in SNNs. For a convolution layer with output spatial size $O\times O$, input and output channels $C_{in}$ and $C_{out}$, and kernel size $k$, the ANN energy is proportional to the dense MAC count:
\begin{equation}
E_{\text{ANN}} = O^2 \cdot C_{in} \cdot C_{out} \cdot k^2 \cdot E_{\text{MAC}} .
\end{equation}
In contrast, spike-driven inference replaces dense MACs with sparse accumulations that occur only when spikes arrive. We therefore scale the operation count by the average firing rate $fr$ measured during inference, and by the effective number of inference steps. When integer-valued training is used and converted to binary spikes by expanding virtual steps, we denote the effective steps as $T\times D$. The resulting SNN energy proxy becomes:
\begin{equation}
E_{\text{SNN}} = (T \times D)\cdot fr \cdot O^2 \cdot C_{in} \cdot C_{out} \cdot k^2 \cdot E_{\text{AC}} .
\end{equation}
We use widely recognised per-operation costs $E_{\text{MAC}}=4.9pJ$ and $E_{\text{AC}}=0.9pJ$ for MAC and AC~\cite{horowitz20141}, respectively, and report total network energy by summing $E$ over all layers. 

\subsection{Experimental Results}
Tab.~\ref{tab:comp_euvp_LSUI} shows the quantitative results on the EUVP and LSUI datasets. Compared with the previous SNN state-of-the-art method VLIF, UIESNN leverages multi-scale information to better capture low-frequency underwater degradations, enabling \textbf{UIESNN} to achieve state-of-the-art performance among SNN methods. Specifically, UIESNN reaches 26.97 dB PSNR and 0.8936 SSIM on EUVP, exceeding VLIF by 1.28 dB and 0.0119. On LSUI, UIESNN attains 24.7346 dB PSNR and 0.8754 SSIM, improving PSNR by 0.56 dB over VLIF and 0.86 dB over ESDNet. We additionally report two ANN-based methods (TACL and UIE-WD) for reference. Fully closing the performance gap between ANNs and SNNs is not the goal of this work. Instead, we focus on advancing SNN-based UIE with strong restoration quality and improved efficiency. Notably, UIESNN outperforms these ANN baselines on both datasets while using only 18.05\% and 42.17\% of their energy consumption (199.33 mJ vs.\ 1104.3 mJ and 472.7 mJ), and it requires 59.1\% of TACL's parameters (16.72M vs.\ 28.29M).

Fig.~\ref{fig:long} presents qualitative results on two benchmark datasets EUVP and LSUI. On EUVP (top row), our method exhibits more accurate global colour restoration and finer-grained appearance details. In the highlighted regions, competing methods tend to produce noticeable colour shifts or oversmoothed textures, which weaken material cues and local contrast. In contrast, our results better preserve subtle texture variations and natural colour transitions, indicating stronger capability in capturing global illumination and colour statistics while maintaining fine details.
On LSUI (bottom row), we observe that ESDNet and VLIF suffer from different degrees of spatial inconsistency, especially around object boundaries and elongated structures, as marked by the red ellipses. A plausible reason is that local perception in spiking networks introduces an inductive bias, so when feature integration relies on local windows or limited receptive fields, small response mismatches between neighbouring regions can accumulate and cause discontinuous structures and unstable geometry. Our method alleviates this issue by strengthening the cross-region feature coupling and reducing the reliance on purely local evidence aggregation, which leads to more coherent contours, smoother structural transitions, and fewer block-like artefacts.

\subsection{Ablation Studies}

\subsubsection{Effectiveness of Each Components}
\begin{table}[t]
\centering
\setlength{\tabcolsep}{0.95mm}
\caption{Ablation study on architectural components.}
\begin{tabular}{cccccc}
\hline
\hline
Model  & FDB & MDA & MPLB & PSNR$\uparrow$   & SSIM$\uparrow$   \\ \hline
(a)     &       &        &         & 25.39  & 0.8643 \\
(b)    &   $\checkmark$     &  &   & 25.45 & 0.8677 \\
(c)    & $\checkmark$ & $\checkmark$ &   & 25.88 & 0.8718 \\ 
Ours    & $\checkmark$ & $\checkmark$ & $\checkmark$ & \textbf{26.97} & \textbf{0.8936} \\
\hline\hline
\end{tabular}
\vspace{-0.7cm}
\label{tab:abl_each_comp}
\end{table}
Table~\ref{tab:abl_each_comp} evaluates the contribution of each component using PSNR and SSIM on EUVP. Compared to the baseline (a), which involves simple spike-driven convolution and tdBN in SRB modules, adding FDB (b) yields a modest improvement, suggesting that frequency decomposition promotes more diverse temporal features. Adding MDA on top of FDB (c) further improves performance by strengthening spatio-temporal and channel interactions. With MPLB enabled, the full model achieves the best results, yielding significant gains of 6.22\%$\uparrow$ / 3.39\%$\uparrow$ (PSNR / SSIM) over (a) and 4.21\%$\uparrow$ / 2.50\%$\uparrow$ over (c). This indicates that the scale-aware representations induced by MPLB are crucial for underwater image enhancement.

\subsubsection{Effectiveness of Different Time Steps and Quantisation Steps}
\begin{table}[t]
\centering
\setlength{\tabcolsep}{3mm}
\caption{Impact of different time steps (T) and quantisation steps (D) on EUVP dataset.}
\begin{tabular}{cccc}
\hline
\hline
\textbf{T × D} & \textbf{Energy ($mJ$)}$\downarrow$ & \textbf{PSNR}$\uparrow$ & \textbf{SSIM}$\uparrow$ \\
\hline
(1 $\times$ 1) & 18.84  &  25.66 & 0.8591 \\
(1 $\times$ 4) & 43.42  &  25.84 & 0.8634 \\
(4 $\times$ 1)  & 52.20 & 26.57 & 0.8902\\
(4 $\times$ 4) \textbf{Ours}  & 199.33 & \textbf{26.97} & \textbf{0.8936} \\
\hline
\hline
\end{tabular}
\vspace{-0.6cm}
\label{tab:TxD}
\end{table}
Tab.~\ref{tab:TxD} evaluates the effects of time steps ($T$) and quantisation steps ($D$) on EUVP. Extending the temporal horizon from $T{=}1$ to $T{=}4$ yields clear gains of 3.55\%$\uparrow$ / 3.62\%$\uparrow$ (PSNR / SSIM), showing that longer time steps facilitate temporal aggregation and better capture complex degradation patterns. Increasing $D$ from 1 to 4 provides smaller gains (0.70\%$\uparrow$ / 0.50\%$\uparrow$) with lower energy cost (43.42 mJ) than increasing $T$ (52.20 mJ). The full setting ($T{=}4$, $D{=}4$) performs best, improving over ($4{\times}1$) by 1.51\%$\uparrow$ / 0.38\%$\uparrow$, but at much higher energy (199.33 mJ vs. 52.20 mJ). Overall, temporal integration contributes most of the gains for modeling underwater degradations in SNNs, while finer quantisation offers limited improvements at a high energy cost.

\section{Conclusion} 
This work studies spiking neural networks for underwater image enhancement (UIE). We explored that, UIE requires correcting spatially extensive, low-frequency degradations such as colour casts and haze-like blur, which are difficult for limited-receptive-field spiking neurons. To solve this, we propose the \textbf{Multi-scale Pooling LIF Block (MPLB)} to expand the receptive field with hierarchical pooling and inject scale-aware context into membrane dynamics, capturing global cues while preserving local details. Building on MPLB, we develop \textbf{UIESNN}, an end-to-end SNN tailored for UIE. Experiments on large-scale real-world underwater datasets show that UIESNN consistently outperforms prior SNN baselines with low energy cost. Future work will extend UIESNN to video enhancement and improve the performance-energy trade-off with more adaptive spiking dynamics.

\bibliographystyle{IEEEtran}
\bibliography{egbib.bib}

\end{document}